\documentclass[conference]{IEEEtran}
\IEEEoverridecommandlockouts
\usepackage{cite}
\usepackage{amsmath,amssymb,amsfonts}
\usepackage{graphicx}
\usepackage{textcomp}
\usepackage{xcolor}
\usepackage{float}
\usepackage{caption}
\usepackage{subcaption}
\usepackage[english]{babel}
\usepackage[utf8]{inputenc}
\usepackage{algorithm}
\usepackage[noend]{algpseudocode}
\usepackage{enumerate}

\usepackage{soul}
\usepackage{xcolor}
\definecolor{reddish}{HTML}{FBB4AE}
\definecolor{blueish}{HTML}{B3CDE3}
\definecolor{magentish}{HTML}{FF00AA}
\definecolor{greenish}{HTML}{a1d99b}

\let\Oldcite\cite
\renewcommand{\cite}[1]{~\Oldcite{#1}}  

\def\BibTeX{{\rm B\kern-.05em{\sc i\kern-.025em b}\kern-.08em
    T\kern-.1667em\lower.7ex\hbox{E}\kern-.125emX}}
\begin{document}

\title{Characterizing the Social Interactions in\\the Artificial Bee Colony Algorithm
\thanks{\IEEEauthorrefmark{1}Corresponding author: mmacedo@biocomplexlab.org.}\thanks{\copyright 20xx IEEE. Personal use of this material is permitted. Permission from IEEE must be obtained for all other uses, in any current or future media, including reprinting/republishing this material for advertising or promotional purposes, creating new collective works, for resale or redistribution to servers or lists, or reuse of any copyrighted component of this work in other works.}
}

\author{\IEEEauthorblockN{Lydia Taw\IEEEauthorrefmark{2}, Nishant Gurrapadi\IEEEauthorrefmark{3}, Mariana Macedo\IEEEauthorrefmark{4}\IEEEauthorrefmark{1}, Marcos Oliveira\IEEEauthorrefmark{5},\\ Diego Pinheiro\IEEEauthorrefmark{6}, Carmelo Bastos-Filho\IEEEauthorrefmark{8}, and Ronaldo Menezes\IEEEauthorrefmark{4}}\vspace{.2cm} 
\IEEEauthorblockA{
\IEEEauthorrefmark{2}Department of Computer Science, George Fox University, USA\\
}
\IEEEauthorblockA{
\IEEEauthorrefmark{3}Department of Computer Science, University of Texas at Dallas, USA\\
}
\IEEEauthorblockA{
\IEEEauthorrefmark{4}BioComplex Laboratory, Department of Computer Science, University of Exeter, UK\\
}
\IEEEauthorblockA{
\IEEEauthorrefmark{5}Computational Social Science, GESIS--Leibniz Institute for the Social Sciences, Germany\\
}
\IEEEauthorblockA{
\IEEEauthorrefmark{6}Department of Internal Medicine, University of
California, Davis, USA\\
}
\IEEEauthorblockA{
\IEEEauthorrefmark{8}Polytechnic School of Pernambuco, University of Pernambuco, Brazil\\}
}

\maketitle


\begin{abstract}
Computational swarm intelligence consists of multiple artificial simple agents exchanging information while exploring a search space. Despite a rich literature in the field, with works improving old approaches and proposing new ones, the mechanism by which complex behavior emerges in these systems is still not well understood. This literature gap hinders the researchers' ability to deal with known problems in swarms intelligence such as premature convergence, and the balance of coordination and diversity among agents. Recent advances in the literature, however, have proposed to study these systems via the network that emerges from the social interactions within the swarm (i.e., the interaction network). In our work, we propose a definition of the interaction network for the Artificial Bee Colony (ABC) algorithm. With our approach, we captured striking idiosyncrasies of the algorithm. We uncovered the different patterns of social interactions that emerge from each type of bee, revealing the importance of the bees variations throughout the iterations of the algorithm. We found that ABC exhibits a dynamic information flow through the use of different bees but lacks continuous coordination between the agents.
\end{abstract}

\begin{IEEEkeywords}
Swarm Intelligence, Network Science, Social Interaction, Artificial Bee Colony
\end{IEEEkeywords}



\section{Introduction}

Since its introduction, Computational Swarm Intelligence has proven to be a valuable approach to numerical optimization and in particular to finding solutions to problems involving high-dimensional search spaces\cite{Bonabeau1999,Kennedy2001,Engelbrecht2006}. Swarm-based algorithms are inspired by the collective behavior exhibited in nature by ants\cite{Dorigo2004}, bees\cite{karaboga2008performance}, fireflies\cite{yang2010firefly}, and fish\cite{bastos2008novel}, to name a few. 
 In swarm-intelligent techniques, multiple simple artificial reactive agents are released onto a limited search space to find the best possible positions\cite{Kennedy2001,Engelbrecht2006}. 
Though several algorithms share this same principle, the community is yet to explain the mechanisms that turn these algorithms intelligent\cite{oliveira2018unveiling}.

Swarm-intelligent algorithms lack \textit{explainability}. We still cannot explain why they work well and---maybe more importantly---why they sometimes do not work\cite{oliveira2018unveiling}.
The field lacks a unified classification of the plethora of algorithms proposed thus far as well as the understanding of their relationships and use cases\cite{oliveira2018unveiling,bonabeau2000inspiration,Sorensen2015}. 
Bonabeau et al. argued that a more systematic comparison of the existing swarm-based algorithms is crucial and could lead to a better understanding of why they work\cite{bonabeau2000inspiration}. After almost two decades since this claim, there has not been much progress made towards a unified framework of swarm intelligence algorithms.
 
Bratton and Blackwell proposed a straightforward way to tackle the problem of understanding swarm algorithms\cite{bratton2007understanding}. They simplified a swarm-based algorithm, namely the Particle Swarm Optimization (PSO)\cite{eberhart1995new}, to study the parts of the system and understand their role in the complex swarm behavior. For this, they removed the randomizing factors from the equations of the algorithm. They demonstrated that by removing components of the algorithm and seeing how the removal affects the performance, we gain insights into what makes PSO an effective method. However, this was still far from showing why social behavior emerges from the simple rules.

A more concrete effort into understanding why swarm-based methods work uses an approach inspired by Network Science proposed by Oliveira et al.\cite{oliveira2018unveiling}. Their approach looks closer at the characteristics of the social interactions within a swarm. This framework can identify the {\it interaction diversity} throughout the iterations and trends towards premature convergence of PSO\cite{complenet2013,oliveira2014towards,Oliveira2015,oliveira2016communication,oliveira2017better}. 

In this paper, we design and analyze the interaction network for the case of Artificial Bee Colony (ABC). We not only show that ABC can be modeled and evaluated by the interaction network approach but also demonstrate the potential of the framework in mapping swarm intelligence techniques to a common space.

The paper is structured as follows: Section \ref{abc} describes the ABC algorithm laying the foundation for understanding our approach. Section \ref{interactiongraph} defines what interaction networks are and why they are relevant to the field of swarm intelligence. In Section \ref{proposal}, we propose how to model the interaction network for the ABC algorithm. Section \ref{experiments} experimental setup applied to our ABC experiments leading to  Section \ref{results} where we discuss the results and the insights obtained from analyzing the social interactions within ABC. Section \ref{conclusions} concludes the paper with some final thoughts regarding directions for future research.

\section{Artificial Bee Colony}\label{abc}


Artificial Bee Colony (ABC) is an optimization algorithm inspired by the foraging behavior of honeybees; it consists of agents (i.e., bees) that change their roles over time\cite{karaboga2005idea}. Each role defines (a) how bees interact with other bees, and (b) how bees search for the solution of a problem. In the algorithm, a \textit{food source}  represents a possible solution to an optimization problem which is evaluated using a fitness function. The roles are defined as follows: 
\begin{enumerate}[(i)]
  \item {\it Employed Bees} exploit food sources to which they are currently assigned (or ``employed'' at) and carry information about the profitability of these food sources. An employed bee interacts with a randomly-chosen employed bee to determine the direction of its movement. 
  
  \item {\it Onlooker Bees} interact with employed bees and are more likely to interact with bees that are employed at the best-known regions of the search space.
  
  \item {\it Scout Bees} are an employed bee that abandoned its food source after exhaustively exploiting its food source.  Scout bees do not interact with other bees and randomly explore the search space seeking for a new food source. Despite the lack of direct interaction, scouts find places that will be occupied by employed bees which in turn will interact with other bees. Hence they lead to future interactions.
\end{enumerate}

 Each cycle of the algorithm consists of bees changing roles, rotating from employed, to onlooker, and then scout. The pseudocode of ABC is described in Algorithm~\ref{alg:ABC}. 
 
 \begin{algorithm}
\caption{Pseudocode for the ABC algorithm.}
\label{alg:ABC}
    \begin{algorithmic}
    \State Initialize the swarm on random positions in the search space
    \While {the stopping criteria is not met}
    \State Place employed bees on the known food sources 
    \State Place onlooker bees on the known food sources
    \State Send scouts to new food sources
    \EndWhile
    \end{algorithmic}
\end{algorithm}
 
 Each employed bee performs a greedy search for a position near to its assigned food source and updates its food source to the newly found position if it has better fitness than the current food source.
Equation~\ref{movement} defines the new position $\vec{v}_{i}$ of the bee $i$: \begin{equation}\label{movement}
v_{i_k}= x_{i_k} + \phi_{i_k}(x_{i_k} - x_{j_k})
\end{equation}
where $x_{i_k}$ is the previous position of the bee $i$ at dimension $k$, $\phi_{i_k}$ is a randomly generated number from a uniform distribution in the interval $[-1, 1]$, and $x_{j_k}$ is the position of a randomly chosen food source $j$ at dimension $k$. 


The onlooker bee is recruited to move towards the food source where another bee is depending on the quality associated with that food source, $p_i$, calculated using Equation~\ref{probability}:
\begin{equation}\label{probability}
p_i = \frac{f(i)}{\sum_{k=1}^{N}f(k)},
\end{equation}
where $f(i)$ is the fitness value of the food source $i$, and $N$ represents the total number of food sources. That is, the bees select a food source using a roulette wheel mechanism that gives better food sources higher chances of being selected by the onlooker bee.
Similar to the employed bees, the onlooker bees perform a greedy search in a nearby position within the search space. 

Once a food source has been fully exploited (following a limiting parameter), the employed bee associated with that food source becomes a scout and selects a new source. 
The scout appears at a random new position and as the algorithm enters the next iteration, where the scout turns into an employed at the newly discovered food source.

\section{Interaction Network}\label{interactiongraph} 

Interactions are at the core of any swarm-based algorithm. The ability of the swarm to coordinate and adapt comes from the rules that define how agents interact among themselves and with the environment\cite{Kennedy2001}. Depending on the inspiration of the algorithm, the interaction may be indirect or direct. Social interactions happen when individuals communicate, but more importantly, they also have the potential to change the individuals\cite{Kennedy2001}.
They are the reason for the emergence of complex behavior within the system, a consequence of agents being able to influence and be influenced by one or several other agents\cite{Bonabeau1999}.

Oliveira et al.\cite{oliveira2018unveiling} introduced the perspective of analyzing swarm intelligence from the social interaction point of view by modeling them using a framework called \emph{interaction network}. In the interaction network $\mathbf{I}$, nodes represent the agents, and the links between the nodes represent the presence (and intensity) of influence between the agents. This network enables the capture of social behavior within the swarm at different points in time during the algorithm execution. 

The authors have used the interaction network to provide the means to understand different aspects of the Particle Swarm Optimization (PSO) such as the search mode and the impact of social dynamics on performance\cite{complenet2013,oliveira2014towards, Oliveira2015,oliveira2016communication,oliveira2017better}. In the case of PSO, Oliveira et al. defined a network $\mathbf{I'}(t)$ at iteration $t$ as an adjacency matrix where each element of the matrix is defined by the presence or the absence of influence between the agent $i$ and $j$ as follows: 
\begin{equation}\label{I}
\mathbf{I'}_{ij}(t) = 
\begin{cases}
    1, & \text{if $i$ interacted with $j$ at the iteration $t$}, \\
    0, & \text{otherwise.}
\end{cases}
\end{equation}
Note that $\mathbf{I'}(t)$ tells us whether two artificial agents in a swarm interacted with each at a given iteration $t$ and neglects the past of the swarm. To keep track of the history of information exchanges, Oliveira et al. defines $\mathbf{I^{t_w}}(t)$ using a time window $t_w$ as the following:
\begin{equation}
\label{iwt}
    \mathbf{I^{t_w}}(t)= \sum_{t_i=t-t_w+1}^{t} \mathbf{I'}(t_i)
\end{equation}
with $t \ge t_w \ge 1$. The time window $t_w$ defines the frequency--recency balance in the analysis\cite{oliveira2014towards}.

The weighted interaction network $\mathbf{I^{t_w}}(t)$ allows us to analyze the history of interactions of each agent in a swarm and determine if any particular agents have significantly more interactions and exerted more influence on the other agents. Moreover, we can analyze different time windows through time to identify the characteristics that define swarm-based techniques.

In order to capture the structure of the information flow within the swarm, Oliveira et al. also proposed to measure how quickly the interaction network can be destroyed by removing the edges from the network\cite{oliveira2018unveiling,oliveira2016communication}. As we remove edges based on their weight, isolated components start to emerge in the network\cite{oliveira2018unveiling}. 
Let $C(w)$ be the number of components in the network $\mathbf{I^{t_w}}$ when we remove each edge $\mathbf{I^{t_w}}_{ij}$ that satisfies $\mathbf{I^{t_w}}_{ij} \leq w$. Note that the maximum value of $C(w)$ is the number of nodes in the network. The area under the curve of $C(w)$, namely $A$, gives us a measure of the pace that the components emerge and can be seen as a measure of diversity in the information flow.
In order to account different networks (i.e., different time windows), Oliveira et al. defined the \textit{Interaction Diversity} (${\rm ID}$) as:
\begin{equation}
{\rm ID}(t) = 1 - \dfrac{1}{|S||T|}\sum_{t'_w\in T}A_{t_w=t'_w}(t),
\end{equation}
where $S$ is the size of the swarm, $A_{t_{w}}$ is the area under the destruction curve of removal edges of $\mathbf{I}$ with time window $t_w \in T$, and $T$ is a set of time windows. With this definition, high ${\rm ID}$ values indicate exploration behavior in swarms, while low ${\rm ID}$ is an indicator of exploitation. 

\section{An Interaction Network for ABC}
\label{proposal}

\begin{figure}
\centering
\includegraphics[width=3.48in]{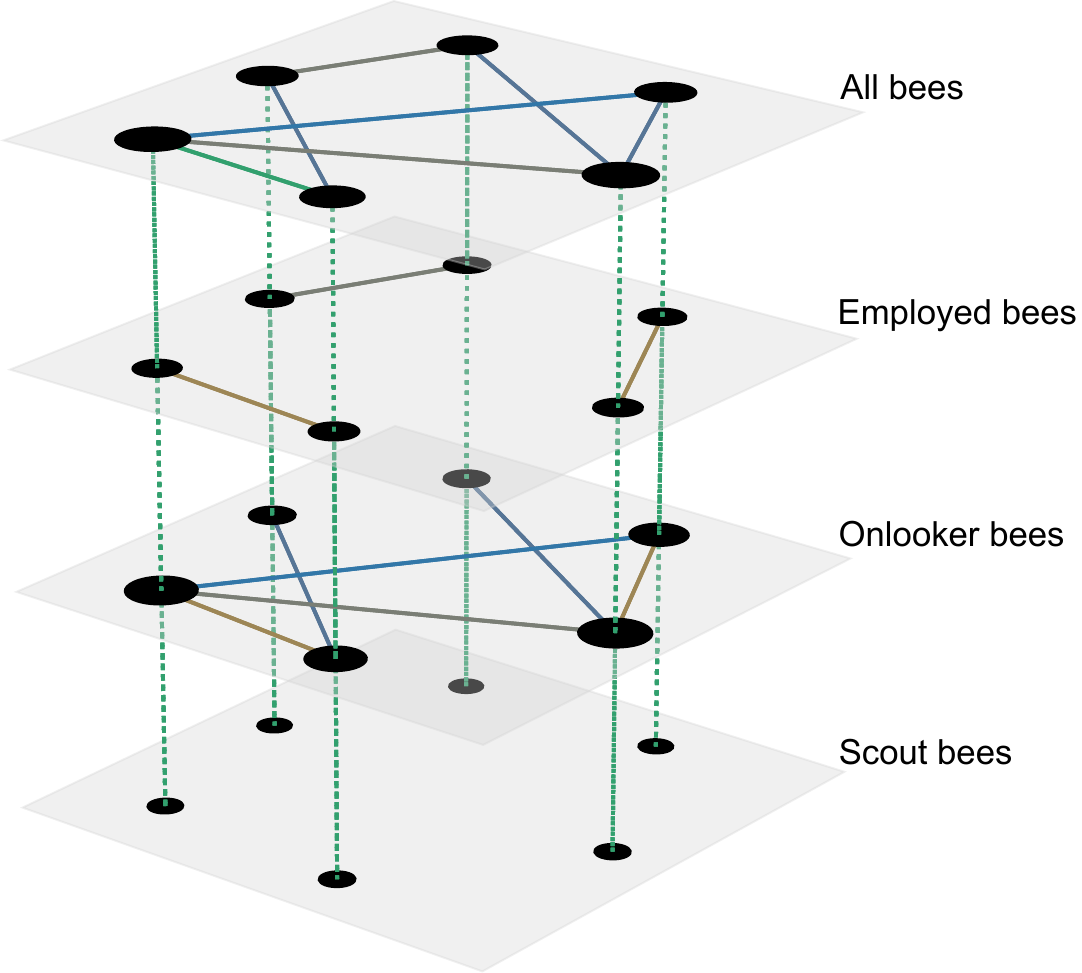}
\caption{
Multi-layer network capturing the social interactions in ABC. 
The nodes represent the bees with their respective degree strength shown by their size. 
The edges colors represent the strength of the link between two bees. The brown color is the weakest link, followed by blue and then green.
We observe that employed bees display more random influences with similar strengths, onlooker bees have a more diverse social interaction network because some bees are more influential, and the scouts bees only influence themselves. The single layer projection of the multi-layer network of all networks is a simple sum of all edges from every pair of nodes. }
\label{fig:layers}
\end{figure}

Looking at ABC from a social interactions perspective, we describe how we model the interaction network for ABC in order to capture what is taking place between the various types of bees and consequently understand the dynamics of the algorithm.
Because each type of bee has a different role in the algorithm, we model an interaction network for each type of bee as a layer in a multi-layer network and use the aggregated projection of the layers as a representation of the entire swarm. Figure~\ref{fig:layers} illustrates this multi-layer representation for ABC. The first layer represents the aggregation of all bee types (a projection of the multi-layer network). The second layer represents the behavior of employed bees who are interacting with each other randomly and with similar interaction strength. The third layer represents onlooker bees and the different strengths between the social interactions which lead to the presence of hubs (bees interacting with many bees) displayed by the size of nodes. The fourth layer represents that scouts are only influencing themselves and not each other. A four-layer model allows capturing the movement and interaction for each type of bee at each iteration of the algorithm. There may be other alternatives to build the model, but this particular one provides a comprehensive way to analyze the abstraction of the core of the social interactions.

\subsection{Interaction Network for the Employed Bees $\mathbf{I^{t_w}_E}$}

In the employed stage, an edge can be drawn between bee $i$ and $j$ if bee $i$ influenced bee $j$ or vice-versa during a successful greedy search. It is important to note that we are representing interactions in such a way that an edge is only drawn if the bee actually moved, indicating the movement was successful. 
\begin{equation}\label{Ie}
\mathbf{I'_E}_{ij}(t) = 
\begin{cases}
    1, & \text{if $i$ moved using information from bee $j$} \\
       & \text{at the iteration $t$}, \\
    0, & \text{otherwise.}
\end{cases}
\end{equation}

\subsection{Interaction Network for the Onlooker Bees $\mathbf{I^{t_w}_O}$}
In the onlooker stage, the bee is recruited to go to a food source and is once again influenced by random bees at that food source. Because bees are more likely to be recruited to promising locations, we would expect to see strong recurrent influences between bees at those food sources leading to networks having hubs (i.e., highly-connected nodes).
\begin{equation}\label{Ie}
\mathbf{I'_O}_{ij}(t) = 
\begin{cases}
    1, & \text{if $i$ moved using information from bee $j$} \\
       & \text{at the iteration $t$}, \\
    0, & \text{otherwise.}
\end{cases}
\end{equation}

\subsection{Interaction Network for the Scout Bees $\mathbf{I^{t_w}_S}$}
An employed bee turns into a scout when the greedy search performed at a food source has been unsuccessful for a predetermined number of trials. The bee jumps to an entirely new, random position on the search space. The scout stage represents the exploration of the algorithm. Excluding it from the interaction network would have resulted in an inaccurate representation of overall influence among the bees. When a bee changes its role to a scout, a self-loop link of influence can be drawn, because the bee is influencing its new position. 
\begin{equation}\label{Ie}
\mathbf{I'_S}_{ij}(t) = 
\begin{cases}
    1, & \text{if $i$ moved using information from bee $j$} \\
       & \text{at the iteration $t$}, \\
    0, & \text{otherwise.}
\end{cases}
\end{equation}

It is worth noticing that in our current model, $\mathbf{I^{t_w}_S}$ is disconnected meaning that it does not contribute directly to the aggregated network. However, we still chose to represent it in the general model because others could choose to capture the indirect influence of scouts in the future. As we mentioned before, scouts lead to future interactions because their random jumps will cause employed and onlooker bee interactions. We discuss this in a little more detail in Section \ref{conclusions}.

\subsection{Aggregation of the Interaction Networks $\mathbf{I^{t_w}_A}$}
We use the same Eq.~\ref{I} and Eq.~\ref{iwt} developed by Oliveira et al., but, in our case, we capture each stage both separately (each layer), and together as a sum of the stages to form the interaction network representing the full run of the ABC algorithm.

\begin{equation}\label{Ie}
\mathbf{I^{t_w}_A} = \mathbf{I^{t_w}_E} + \mathbf{I^{t_w}_O} + \mathbf{I^{t_w}_S}
\end{equation}


Note that this may not be the only way of building an interaction network for ABC, but we believe that our model demonstrates the network can capture some idiosyncrasies of ABC which suffices for understanding the algorithm.

\section{Experimental Setup}\label{experiments}

We selected the Rastrigin function to analyze and understand the behavior of bees in ABC. This multimodal function is widely applied as a benchmark of optimization algorithms. Because of its many peaks, it is a hard function to solve with a high number of dimensions. The optimal solution for the Rastrigin function is at $\vec{x} = \vec{0}$ with $F(\vec{x}) = 0$. To determine whether the complexity of Rastrigin impacts the behavior bees in ABC, we tested it with two different configuration setups: using 100 and 1,000 dimensions. It is important to highlight that the study in this unique benchmark does not allow to analyze the behavior of the swarm in all situations, but enable to validate the proposal as a tool to analyze artificial bee colony algorithms. 

All bees were spread uniformly in the search space at the beginning of each of the 30 executions. We used $50$ bees in our experiments.
Initially, we run the experiments with 200,000 evaluations per execution. However, it became evident that more time was necessary in order for patterns to be detected. Specifically, we needed to detect if the behavior found at the beginning of the execution would happen again later in the execution. Consequently, the number of evaluations per execution was increased to 1,000,000 resulting in around 9,950 iterations per execution. Because the scouting stage results in a non-deterministic number of evaluations, the exact number of iterations varies slightly between executions.

In our results, we aim to show that employed and onlooker bees display different behaviors because onlookers are more likely to go to better regions. Also, we want to show that each bee has the opportunity to influence the swarm at different points in the execution, hence the necessity for having all three mechanisms modeled. To demonstrate this necessity, we evaluate the temporal dynamic behavior of each network. We also show that the aggregation of all bees into one network hides the peculiarities of each type of bee, but reveals the overall effects of the bees working together in each iteration. 


\section{Results}\label{results}

\begin{figure}[b!]
\centering
\includegraphics[width=\columnwidth]{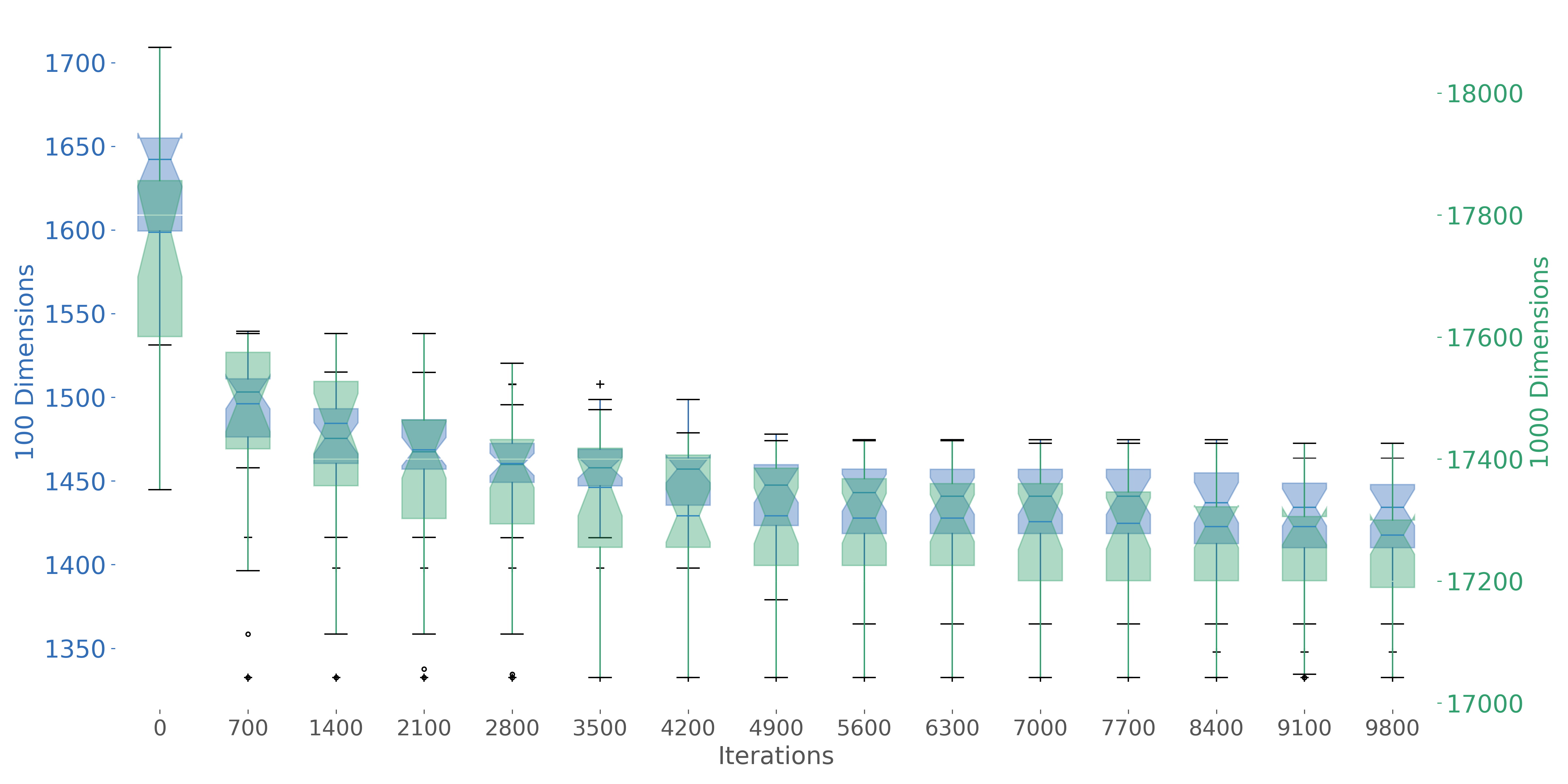}
\caption{Evolution of the fitness as a function of the number of iterations with around 9500 iterations. The double y-axis represents the value of the fitness at each execution time. This compares the behavior of two experiments run with two different configurations. The blue indicates the experiment run with 100 dimensions, and the green indicates 1000 dimensions.}
\label{fig:fitness_100D}
\end{figure}

We first look at the evolution of the fitness value through the interactions. 
Figure \ref{fig:fitness_100D} shows the executions for 100 and 1,000 dimensions. We noticed that the evolution of fitness varied among the executions. All the executions converged prematurely as the best solution would be zero. Some simulations converged before 3,000 iterations, others would continue improving through more iterations. The influence between bees dynamically changes over time due to the nature of the algorithm, so we cannot observe a complete stagnation of the swarm on the search space even though we notice a fitness stagnation. The scout stage allows for new areas to be explored, but ABC does not have mechanisms to prevent bees from exploiting areas similar to their old food sources nor remembers multiple good regions explored in the past.

\begin{figure}[b!]
\centering
\includegraphics[width=\columnwidth]{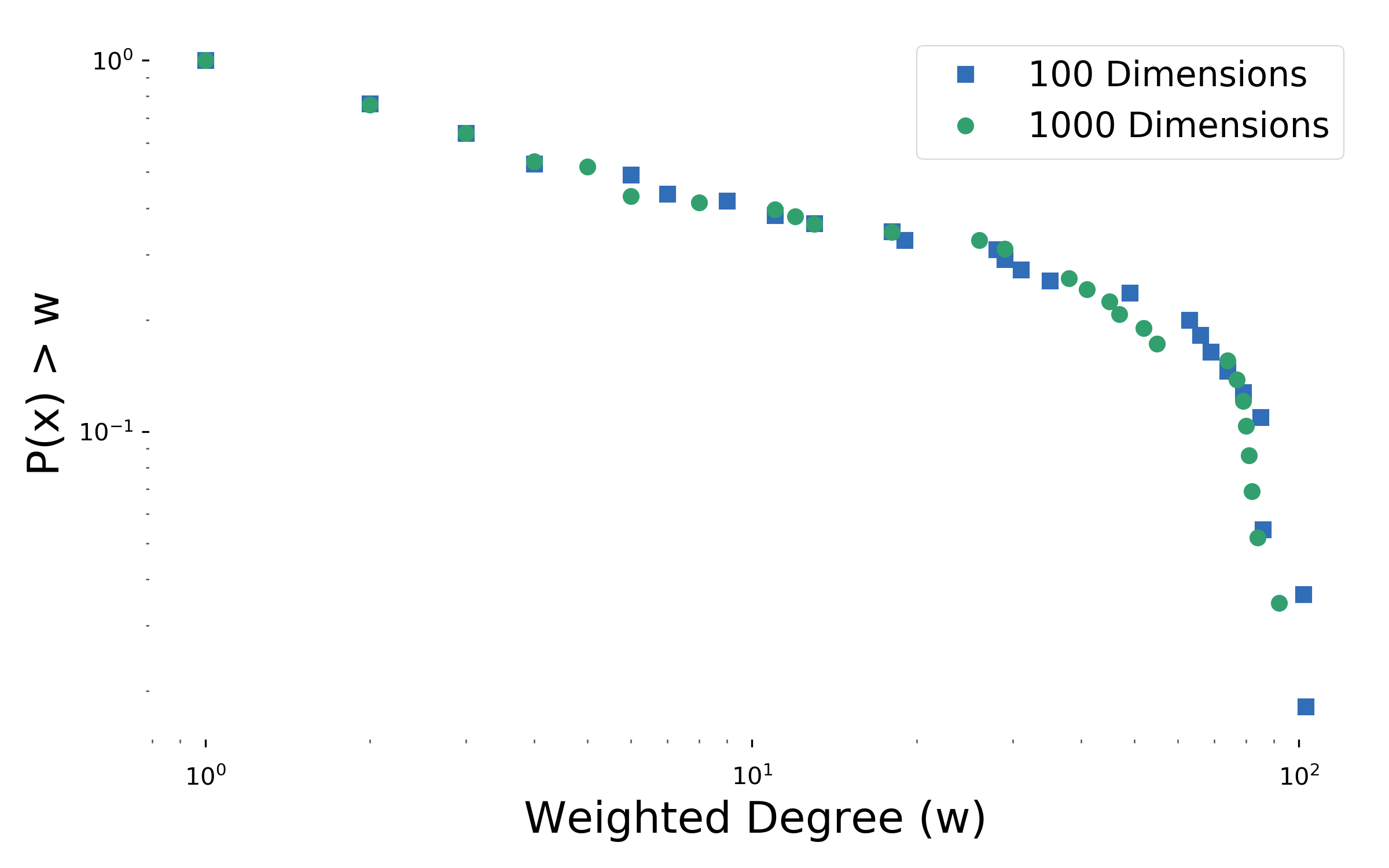}
\caption{Complementary Cumulative Distribution Function (CCDF) of the final interaction network of 100 and 1,000 dimensions. Note that there are very few nodes with a high weighted degree (hubs) Note that the plot is on a log-log scale..
}
\label{fig:ccdf_dimensions}
\end{figure}

Figure~\ref{fig:ccdf_dimensions} depicts the degree distribution of the interaction network at the last iteration. If the bees were interacting randomly, we would see a distribution with a well-behaved expected average degree (normal distribution). Instead, we observe a slow decay that allows bees to attain a high weighted degree---these bees are influential in the swarm. Moreover, the number of dimensions does not change the behavior expected from the interaction network, so we detect similar characteristics from both experiments. 

\begin{figure}
\centering
\includegraphics[width=\columnwidth]{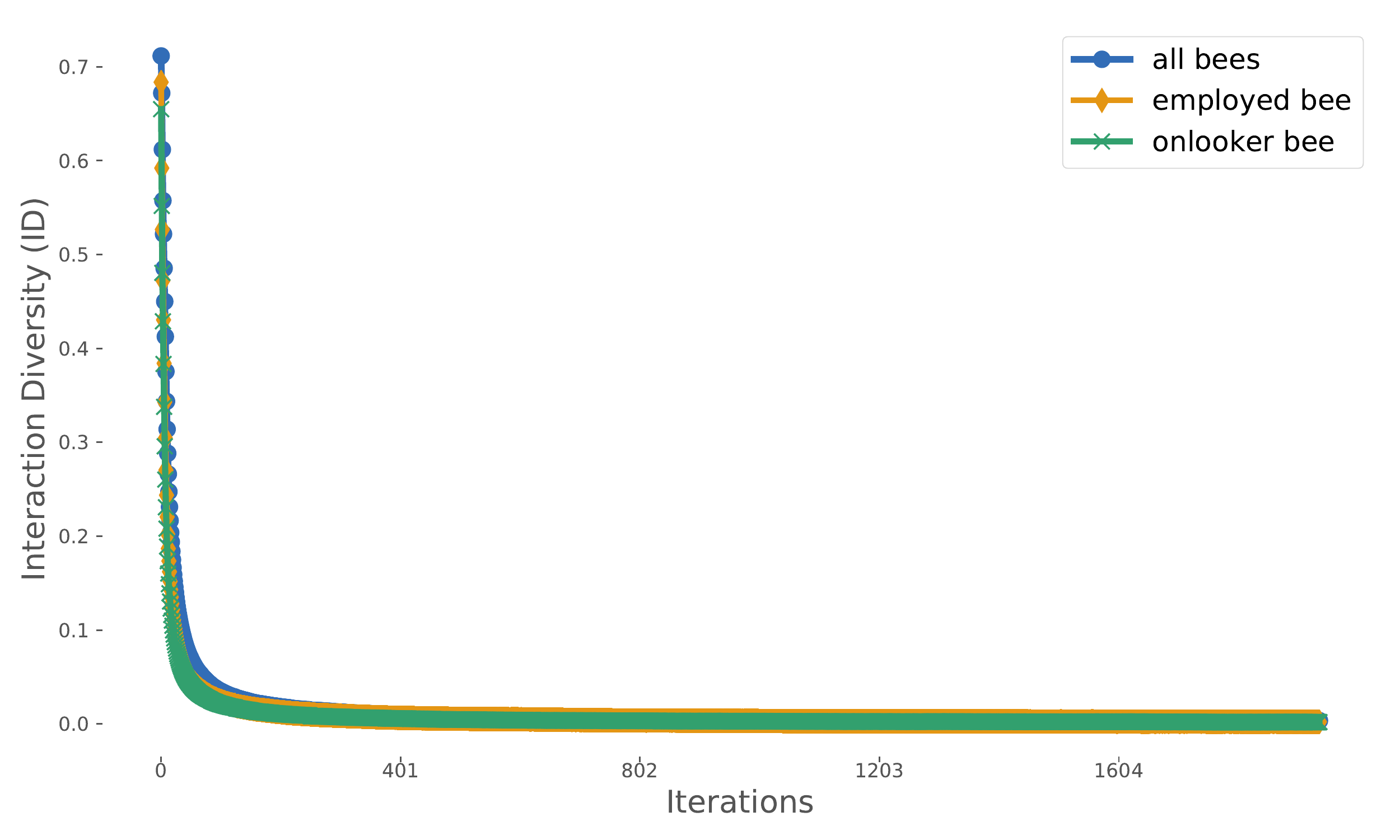}
\caption{Interaction Diversity evolution through the iterations performed by ABC on the Rastrigin function (100 dimensions) shows a smooth decrease in the first 1,000 iterations indicating steady exploitation.}
\label{fig:cd_100D}
\end{figure}

\begin{figure}
\centering
\includegraphics[width=\columnwidth]{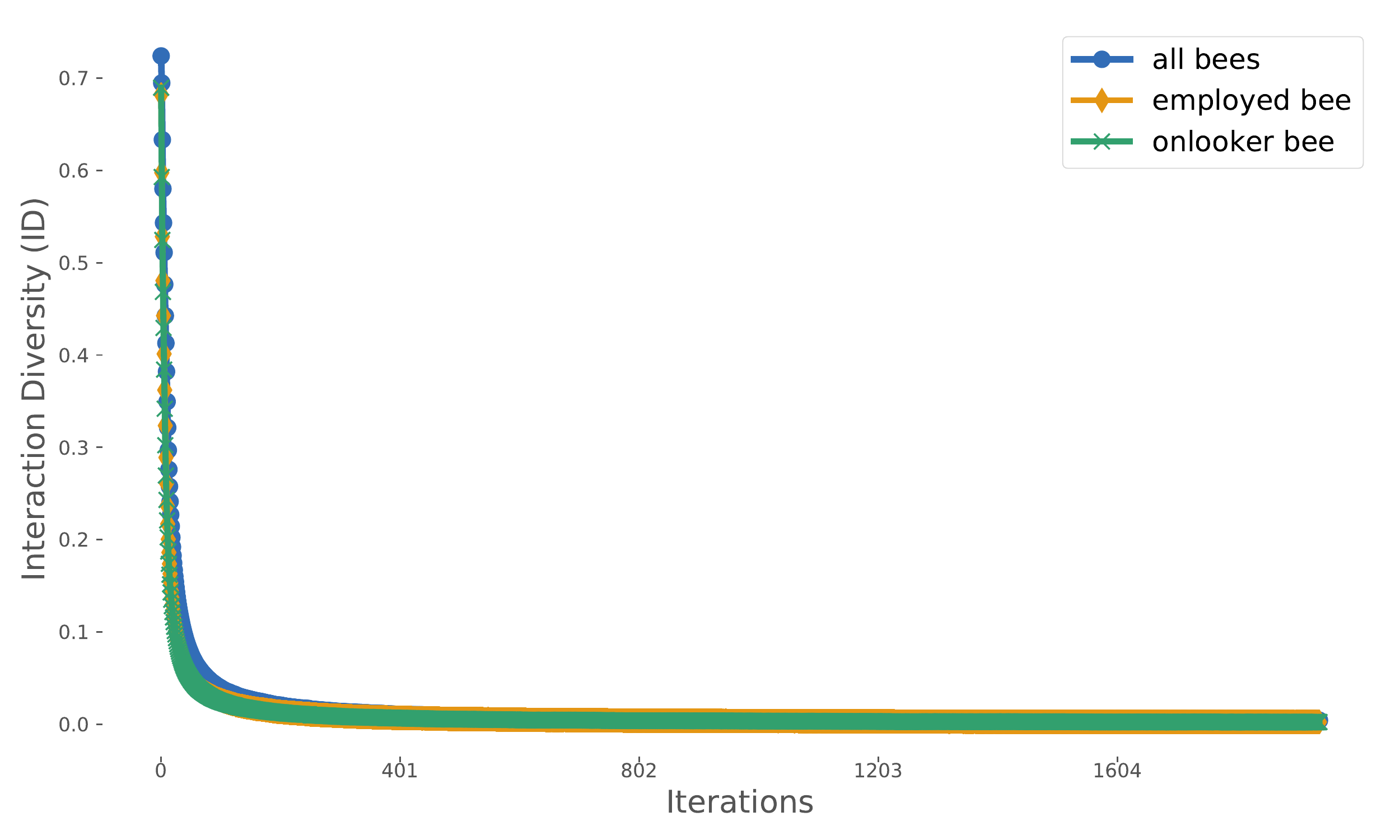}
\caption{Interaction Diversity evolution through iterations performed by ABC on the Rastrigin function (1,000 dimensions). We observe a behavior similar to the case with 100 dimensions.}
\label{fig:cd_1000D}
\end{figure}

Considering the communication flow in the interaction networks, we analyzed the evolution of Interaction Diversity (ID) from 100 and 1,000 dimensions perspectives in Figures \ref{fig:cd_100D} and \ref{fig:cd_1000D} using a cumulative time window of one iteration. Looking at both images, we notice that the difficulty of the function being optimized does not impact the diversity of the swarm, so we conclude that the bees interact similarly independent of the number of dimensions.

The interaction diversity of ABC indicates that exploitation is the main focus of the algorithm, and the little exploration from the lack of the coordination perspective can corroborate ABC being stagnated from the fitness perspective. Even though the scouts bring to ABC new search space areas, they do not have the possibility of high exploration because the scouts have no memory to avoid previously-explored locations or strategies to better spread on the search space.

In the first 100 or so iterations, there is a sharp decrease in Interaction Diversity. This indicates that the swarm is exploiting more over time. Even though the scouts affect the search space, interaction diversity is not impacted. Consequently, the scouts seem to be not relevant to the interaction network point of view.

If we calculate the Interaction Diversity using a time window of 5 iterations, we identify that the diversity does not change much over time because the rules also do not change; we only identify a tiny difference from the values (approx 0.025). The creation of adaptive versions of ABC could be the solution to recover the behavior found at the beginning of the execution and to improve the lack of exploration found at later stages of the execution. However, as ABC does not utilize previous information as much as PSO \cite{eberhart1995new} and FSS \cite{bastos2009fish}, the convergence is the real limitation of the algorithm.

\begin{figure}
\centering
\includegraphics[width=\columnwidth]{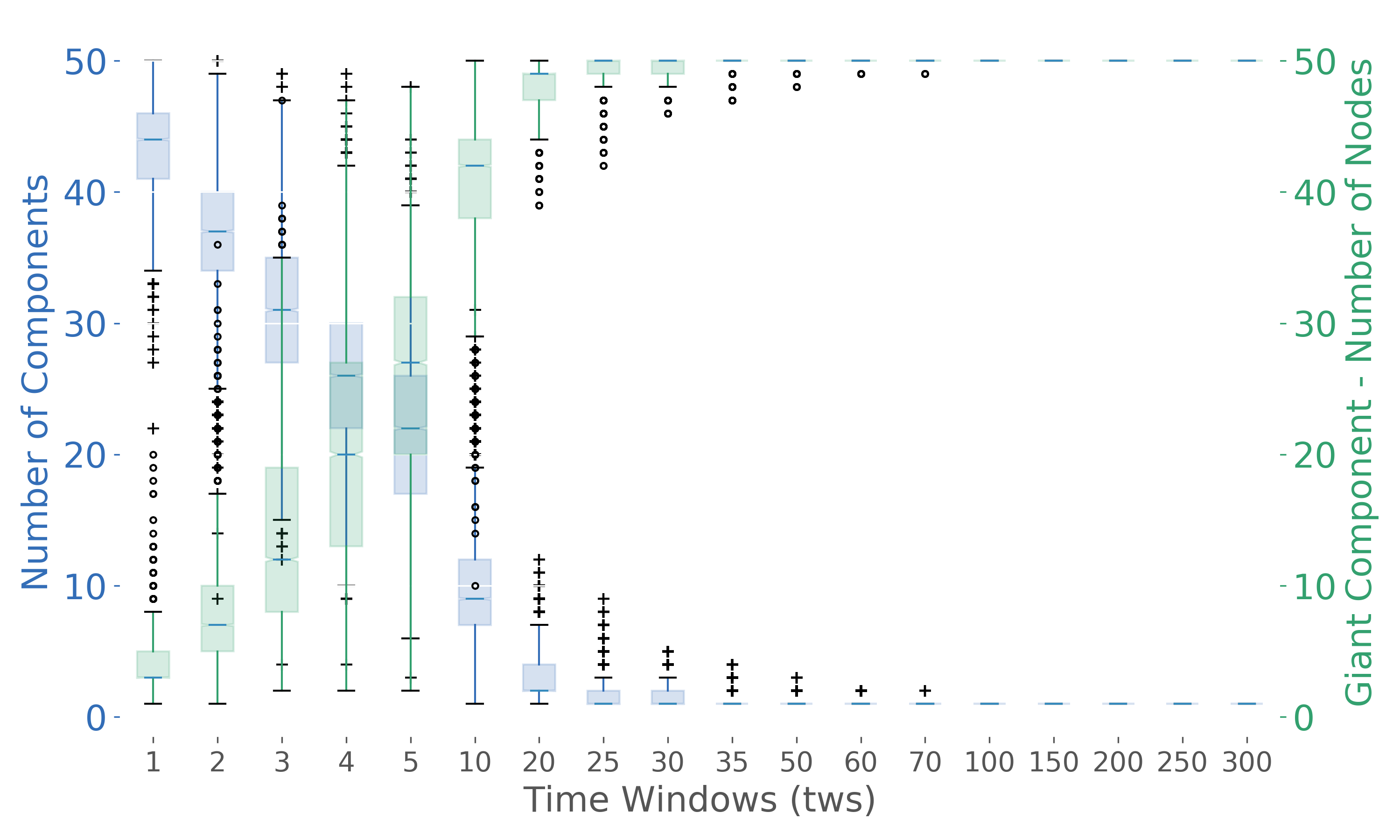}
\caption{Number of components in the interaction network and the number of nodes in the giant component with increasing sizes of sliding time windows (1,000 dimensions).}
\label{fig:all_bees_dim_1000_eval_1000000_sim_15_number_components_giant_component_nodes}
\end{figure}

Figure \ref{fig:all_bees_dim_1000_eval_1000000_sim_15_number_components_giant_component_nodes} shows the number of nodes in the giant component of interaction graph from the best execution using 1,000 dimensions; it also shows how it changes depending on the size of the time window used. The giant component represents the biggest sub-swarm with the most number of nodes concerning the other components. Presumably, the number of components is the same as the number of bees in the swarm upon initialization as no interactions have occurred yet. Throughout the iterations, as the bees start interacting with each other, the number of components decreases as new connections are established and lost. Similarly, the number of nodes in the giant component increases as the time window increases as a result of the cumulative interactions among the bees.

When the time windows are less than 10, we observe that the ABC works on separate sub-swarms. This excess of diversity causes a lack of coordination among the bees, so even though they help each other to move around some food source, they also do not influence each other in a coordinated way. If we analyze what is happening on the search space, they look like they are coordinated in some food sources, but what they are doing is being selfish regarding the neighbors.  

At the same time window point, we see that despite the existence of several components, the giant component seems to be large compared to the other components. The scouts provoke the presence of more sub-swarms. However, the onlookers display a high impact through time enhancing the relationship between some particular bees, and the employed bees make the bees widely connected because they rarely interact with the same bee in the same time window. They are the mechanism by which the swarm tries to coordinate and make bees go to better regions. Without the onlooker role, the convergence of the algorithm would be impacted because they would lack coordination leading to a near random behavior. 

\begin{figure}
\centering
\includegraphics[width=\columnwidth]{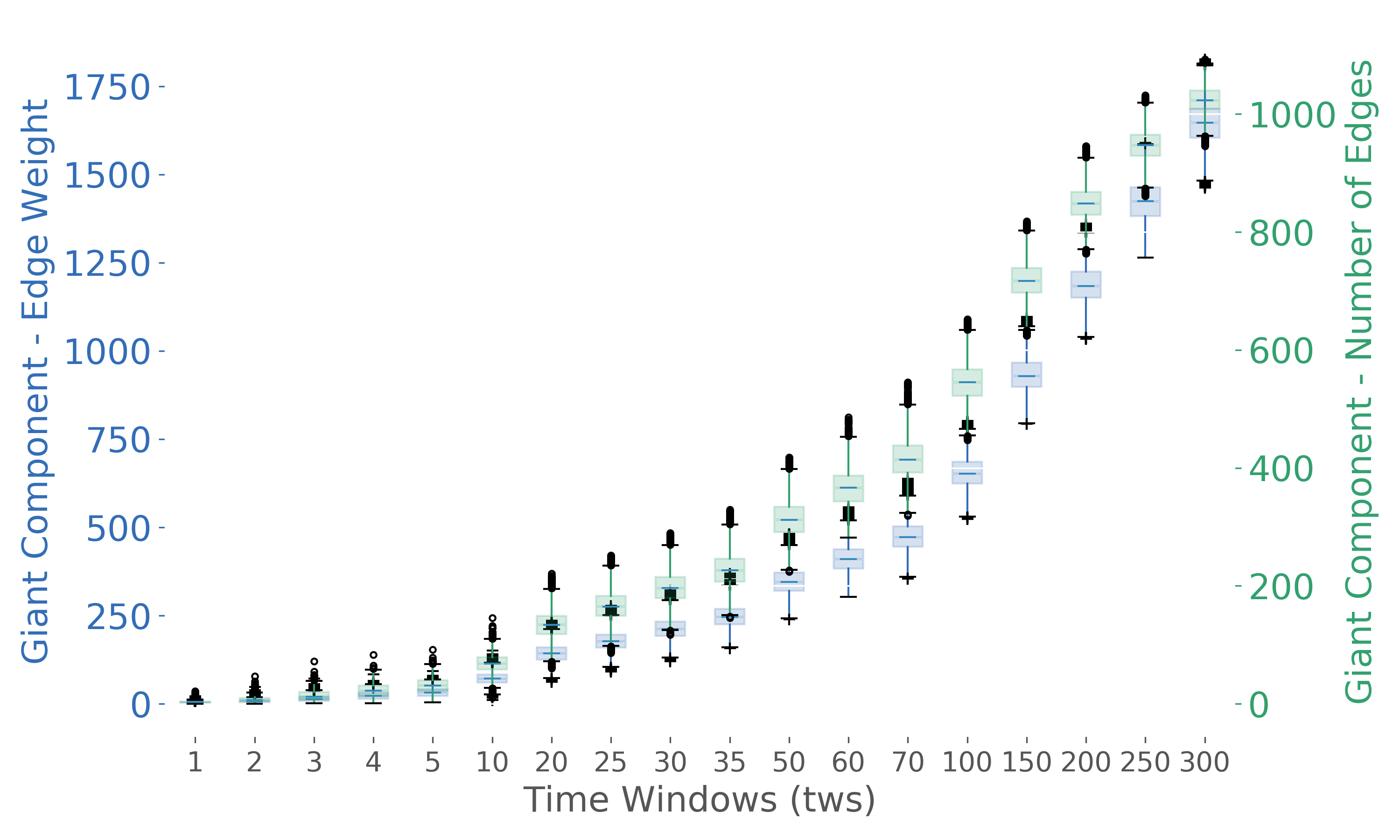}
\caption{Number of edges and sum of the edge weights within the giant component as we vary the size of the time windows (1,000 dimensions). 
}
\label{fig:all_bees_dim_1000_eval_1000000_sim_15_giant_component_edge_weight_giant_component_edges}
\end{figure}

Figure \ref{fig:all_bees_dim_1000_eval_1000000_sim_15_giant_component_edge_weight_giant_component_edges} represents the value of sum of all edge weights and number of edges while increasing the time window from the aggregated network in ABC. We display both because they provide a better idea of the number of edges and the total influence exerted by them. 
The influence of bees (edge weight) and the number of connections (number of edges) are small when analyzing small time windows. Moreover, the increase of the standard deviation shows that the influence fluctuates more throughout the iterations when analyzing bigger time windows. One thing to point out is that even though the greedy movement constrains the bees to move only when successful, the influence by time window remains generally small. Several bees move and are unsuccessful which demonstrates the waste of computational time.  In the first 10 iterations, the bees easily influence each other and move successfully, but after about 10 iterations the movement is less likely to take place.

\begin{figure*}[!h]
\centering
\centering
	\begin{subfigure}[h]{0.49\columnwidth}
		\centering
		\includegraphics[width=\textwidth]{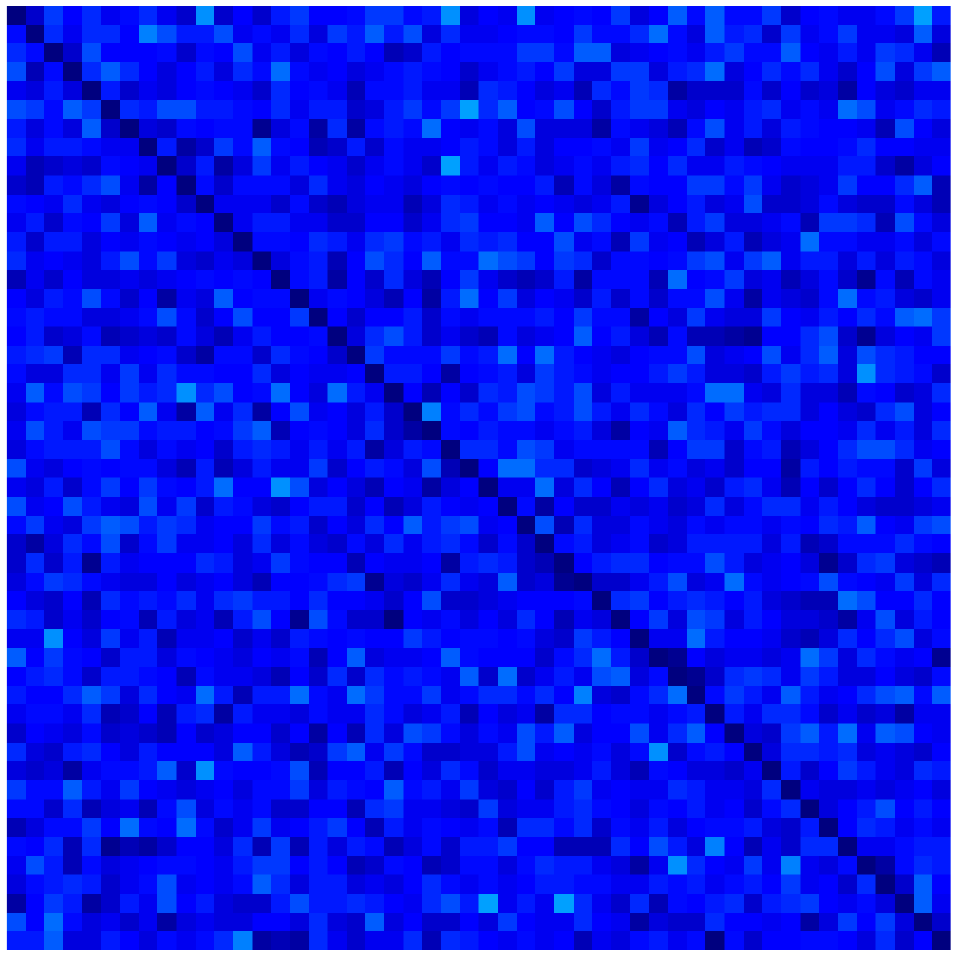}
		\caption{Employed bee}\label{fig:1000D_heatmap_employee_from_it_0_until_4972}
	\end{subfigure}
	\hfill
	\begin{subfigure}[h]{0.49\columnwidth}
		\centering
		\includegraphics[width=\textwidth]{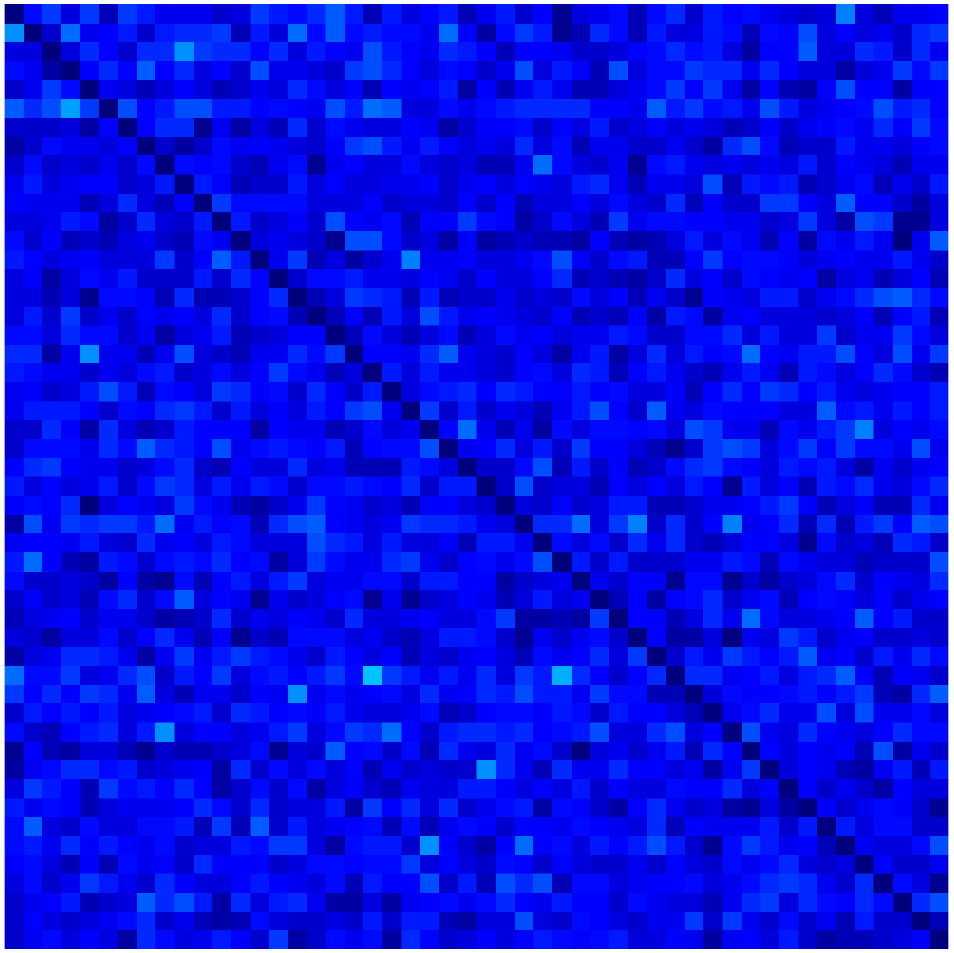}
		\caption{Onlooker bee}\label{fig:1000D_heatmap_onlooker_from_it_0_until_4972}
	\end{subfigure}
	\hfill
	\begin{subfigure}[h]{0.49\columnwidth}
		\centering
		\includegraphics[width=\textwidth]{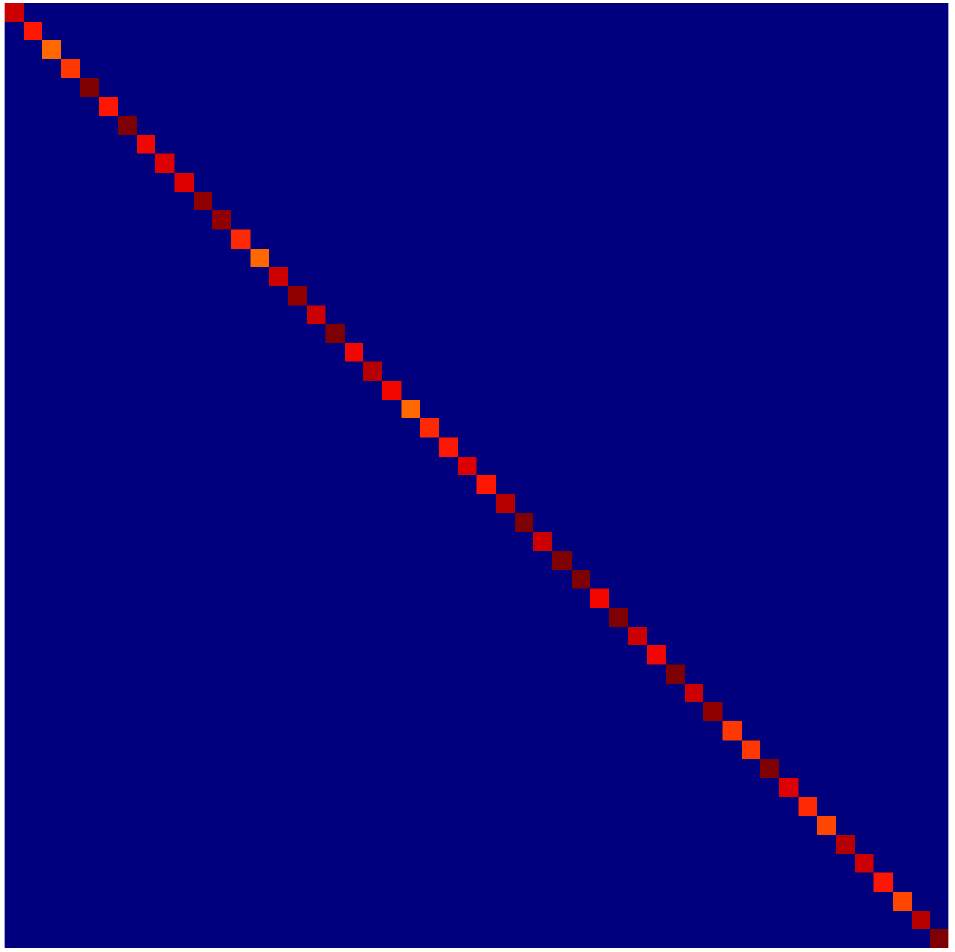}
		\caption{Scouts bee}\label{fig:1000D_heatmap_scouts_from_it_0_until_4972}
	\end{subfigure}
	\hfill
	\begin{subfigure}[h]{0.49\columnwidth}
		\centering
		\includegraphics[width=\textwidth]{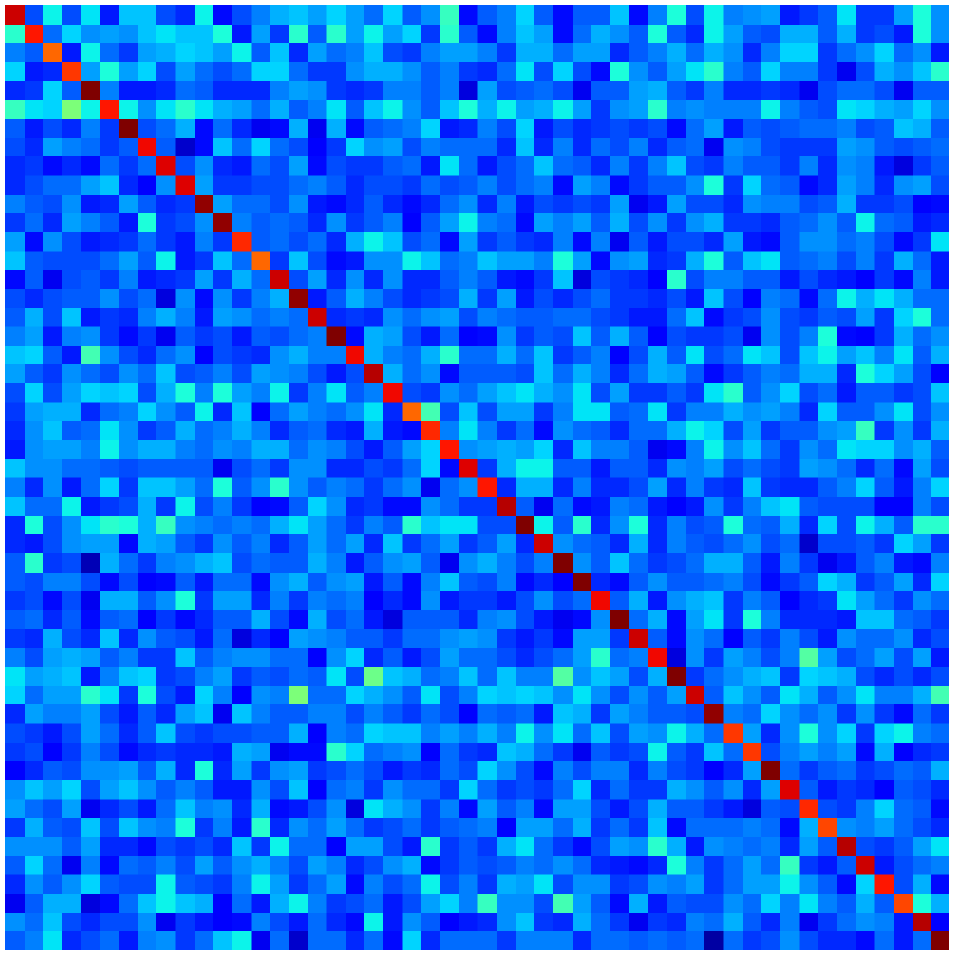}
		\caption{All bees}\label{fig:1000D_heatmap_all_bees_from_it_0_until_4972}
	\end{subfigure}
\caption{
Interaction Network of the middle of iterations for each bee. The degree of influence is indicated on a color scale of dark blue to dark red, with the dark blue indicating a low level of interaction and dark red a high level of interaction. As expected, employed and onlooker bees never influence themselves, and the scout bees only influence themselves.
}
\label{fig:interactiongrapheachbee}
\end{figure*}

\begin{figure}[!h]
	\centering
	\begin{subfigure}[h]{0.49\columnwidth}
		\centering
		\includegraphics[width=\textwidth]{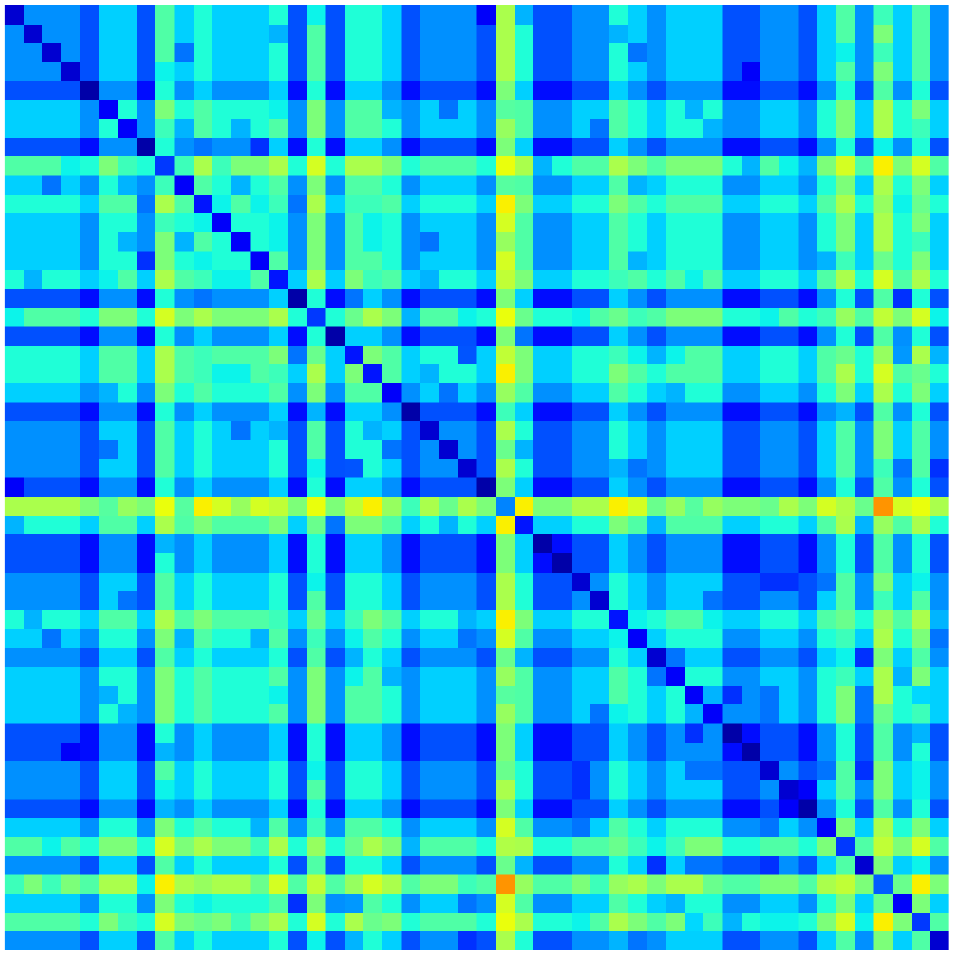}
		\caption{Employed bee - 100D}\label{fig:100D_heatmap_order_employee_from_it_0_until_2485_max_55}
	\end{subfigure}
	\hfill
	\begin{subfigure}[h]{0.49\columnwidth}
		\centering
		\includegraphics[width=\textwidth]{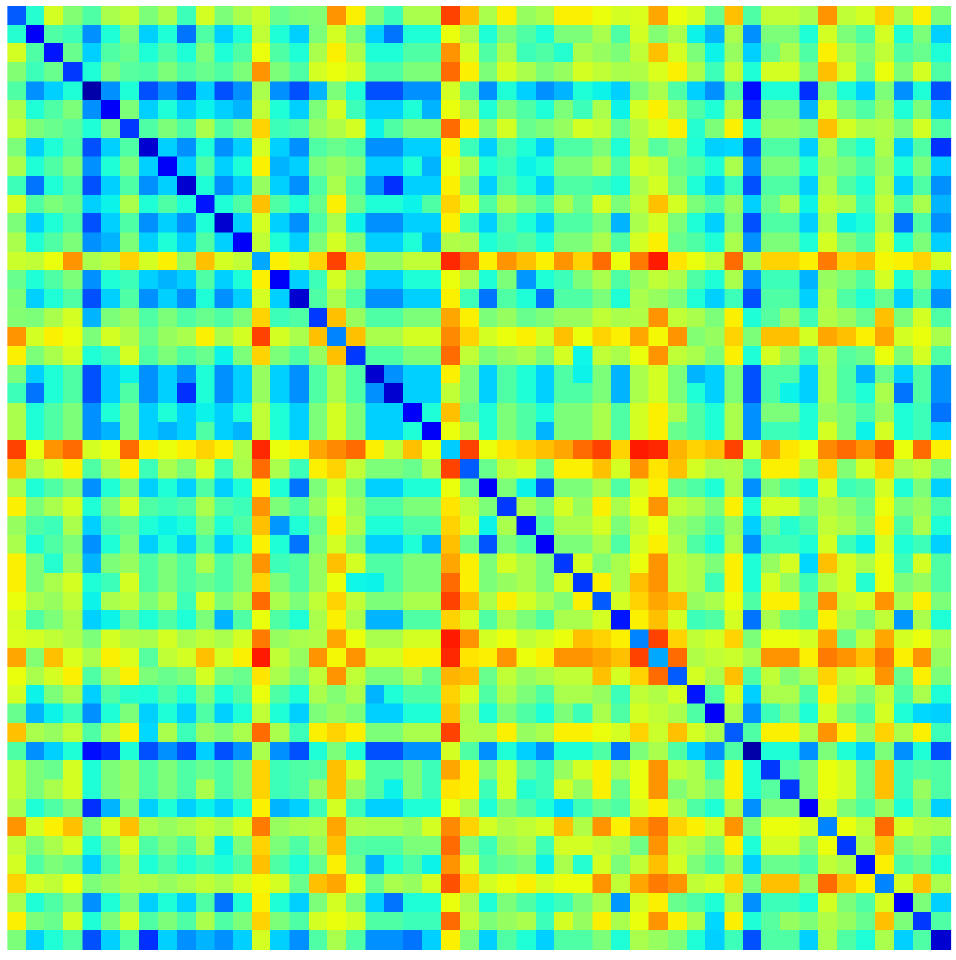}
		\caption{Onlooker bee - 100D}\label{fig:100D_heatmap_order_onlooker_from_it_0_until_2485_max_55}
	\end{subfigure}
	\begin{subfigure}[h]{0.49\columnwidth}
		\centering
		\includegraphics[width=\textwidth]{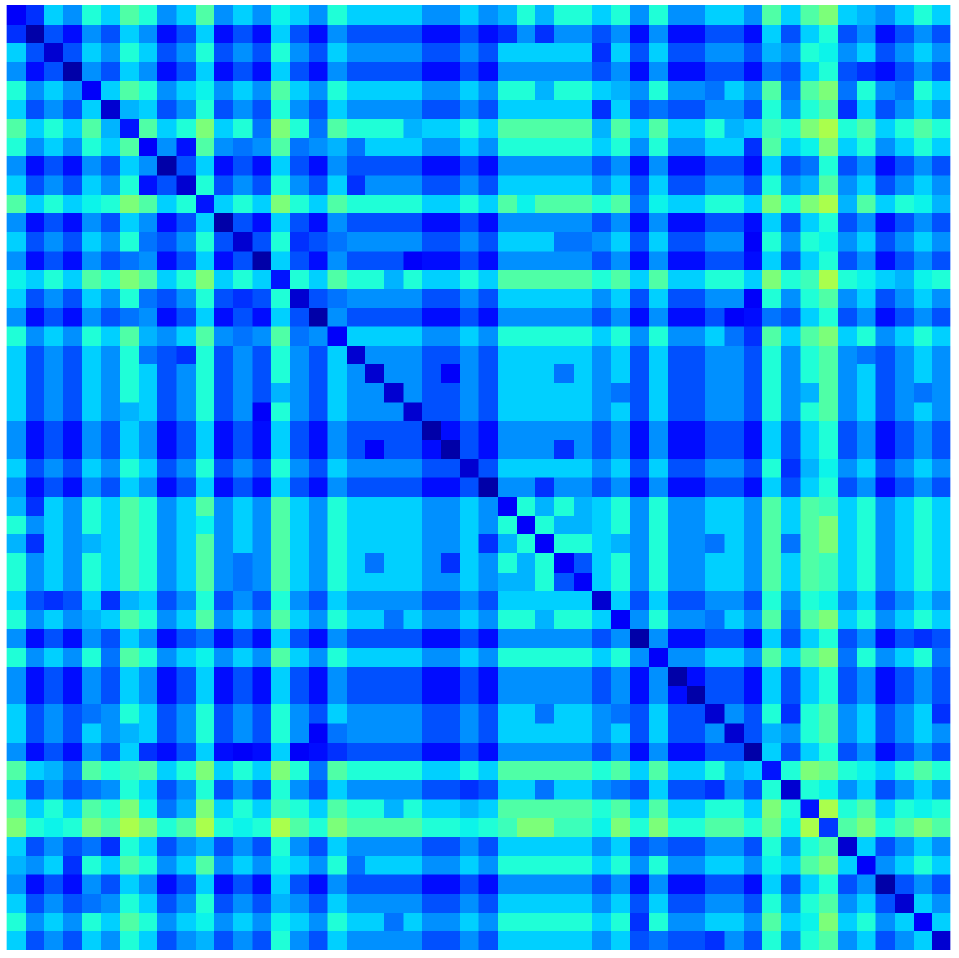}
		\caption{Employed bee - 1000D}\label{fig:1000D_heatmap_order_employee_from_it_0_until_2486_max_55}
	\end{subfigure}
	\hfill
    \begin{subfigure}[h]{0.49\columnwidth}
		\centering
		\includegraphics[width=\textwidth]{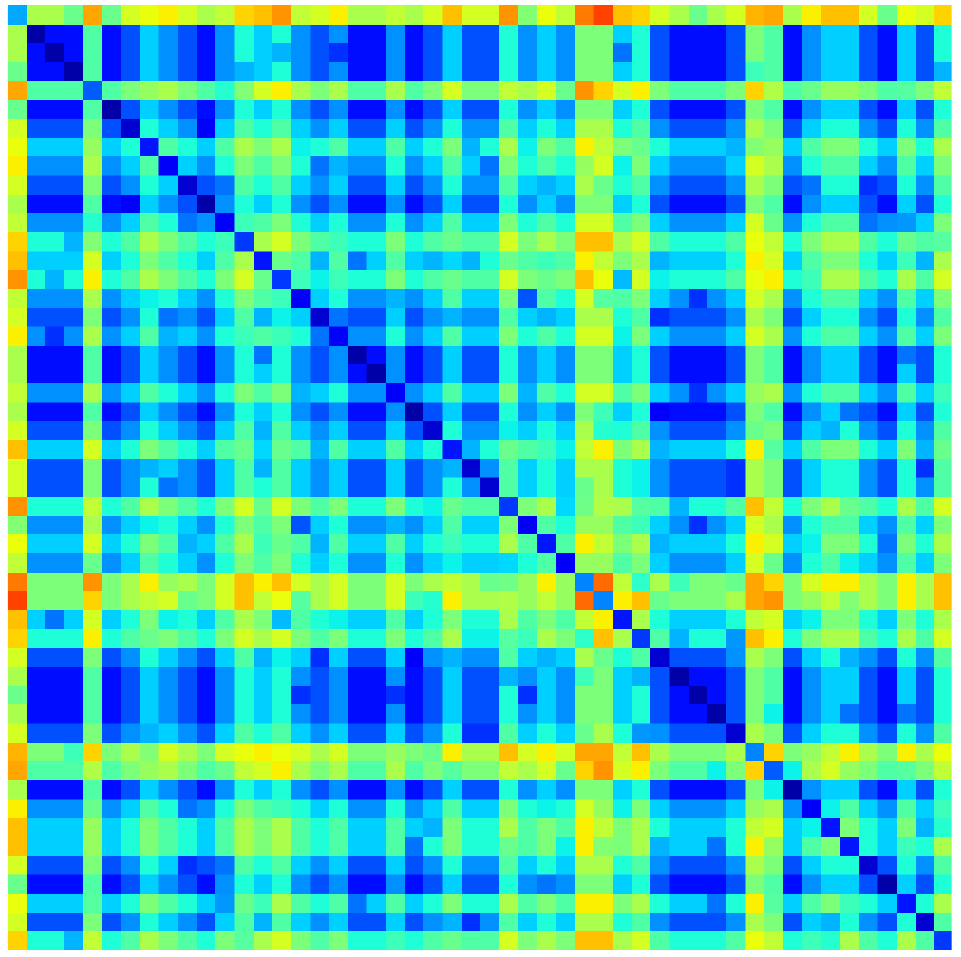}
		\caption{Onlooker bee - 1000D}\label{fig:1000D_heatmap_order_onlooker_from_it_0_until_2486_max_55}		
	\end{subfigure}
	\caption{Interaction network from the employed and onlooker bees for 100 and 1,000 dimensions. The color scheme represents the strength of influence between bees, from blue, the weakest influence until red,  the strongest influence.}\label{fig:comparisonemployeeonlooker}
\end{figure}

In Figure \ref{fig:interactiongrapheachbee}, we show the interaction network (as an adjacency matrix) of each bee type. The interaction network of the employed bees is similar to the onlooker because they differ only in strength and not in the pattern of social interactions. The scout interaction network has only diagonal values, which is precisely what was expected due to the nature of the scouting stage. 
The cumulative networks are scale-free, which means the influence is spread dynamically among the bees, but some bees are hubs (influencers in the swarm). To demonstrate the difference between employed and onlookers, we provide Figure \ref{fig:comparisonemployeeonlooker}; we diminished the color scale, allowing us to observe the intensity of influence between onlookers and employeds. 
The employed stage displays green lines which indicate the presence of hubs in the network because some employed bees start in good regions, while others start in bad regions. The onlooker stage displays green, yellow, and red which represents a higher level of influence than the employed bees. This is because onlooker bees have a higher probability of going towards good regions due to the nature of the algorithm allowing them to have more movement. Furthermore, the bees are more likely to choose the same good locations. 
These heat maps allow us to differentiate between the nuances of influence between employed and onlooker. We can also see that some bees are hubs in both stages, and the onlooker stage is more effective in the greedy search than the employed stage. This is expected because the best food sources always attract more onlooker bees. 

\begin{figure}[!h]
\centering
	\begin{subfigure}[h]{0.49\columnwidth}
		\centering
		\includegraphics[width=\textwidth]{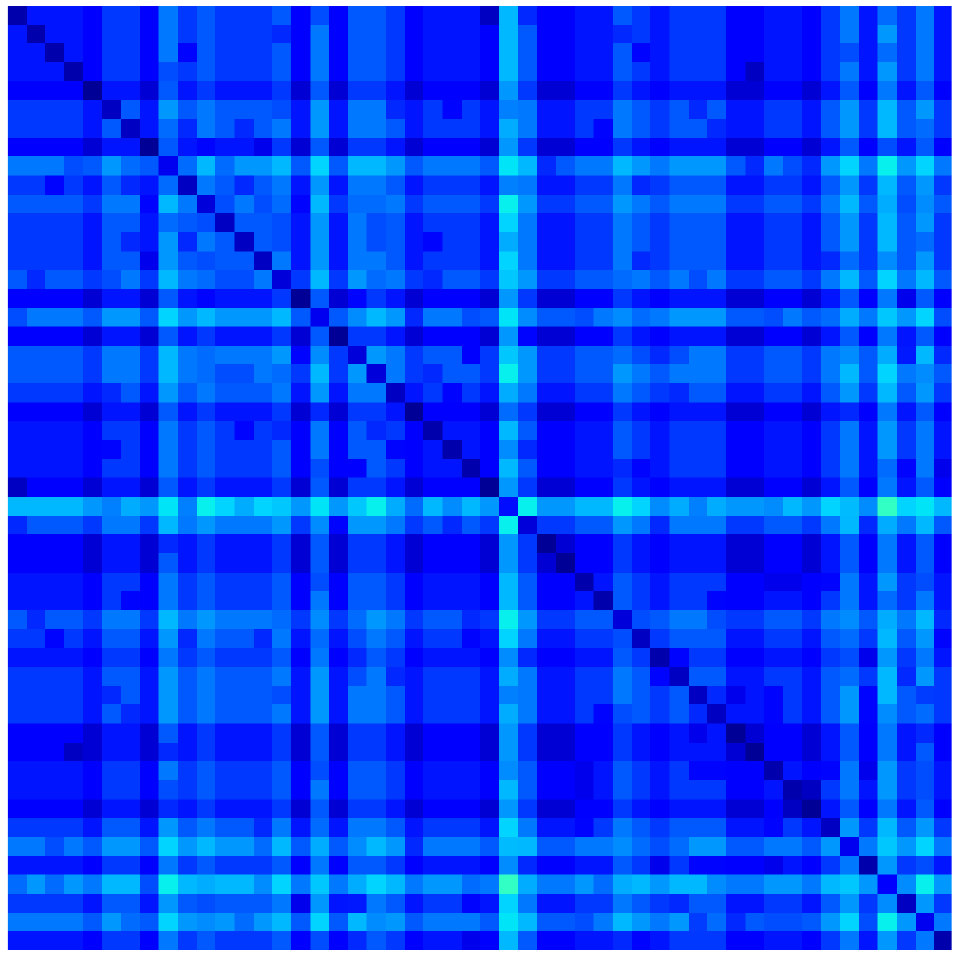}
		\caption{Employed network up to a quarter through the simulation with 100 dimensions}\label{fig:100D_heatmap_order_employee_from_it_0_until_2485}
	\end{subfigure}
	\hfill
	\begin{subfigure}[h]{0.49\columnwidth}
		\centering
		\includegraphics[width=\textwidth]{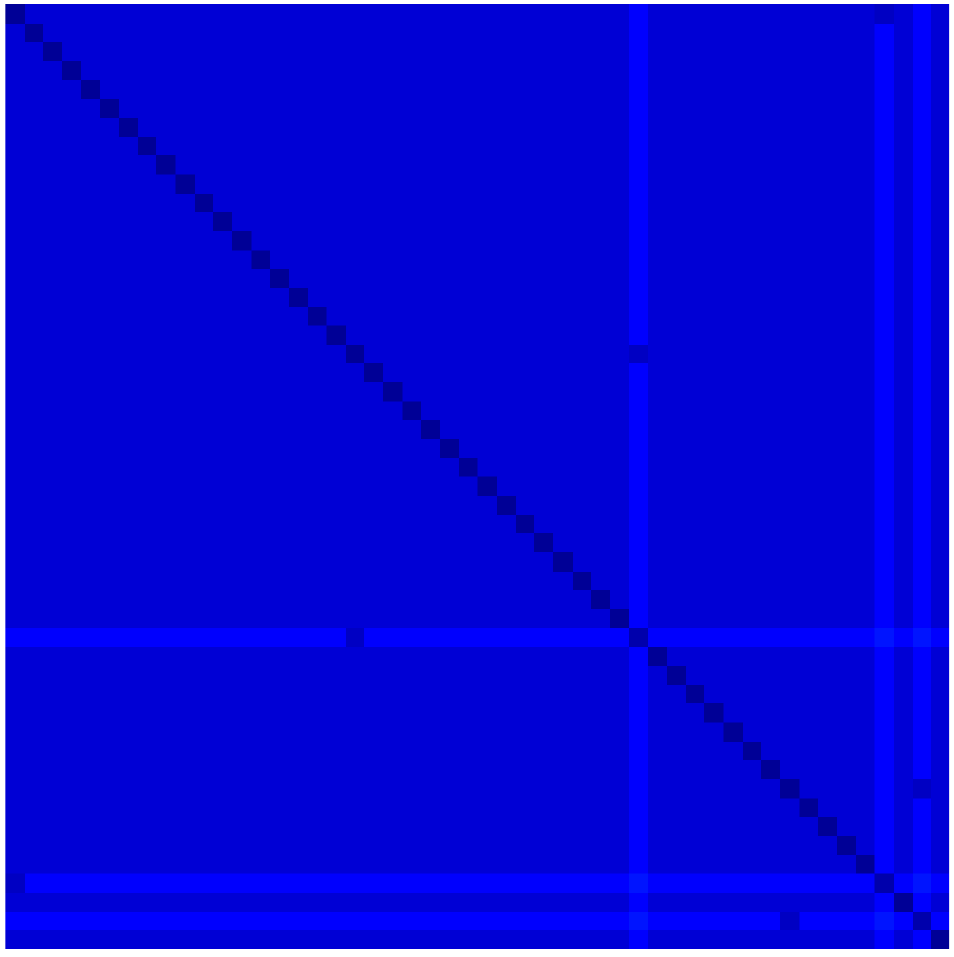}
		\caption{Employed network up to halfway of the simulation with 100 dimensions}\label{fig:100D_heatmap_order_employee_from_it_0_until_4970}
	\end{subfigure}
    \begin{subfigure}[h]{0.49\columnwidth}
		\centering
		\includegraphics[width=\textwidth]{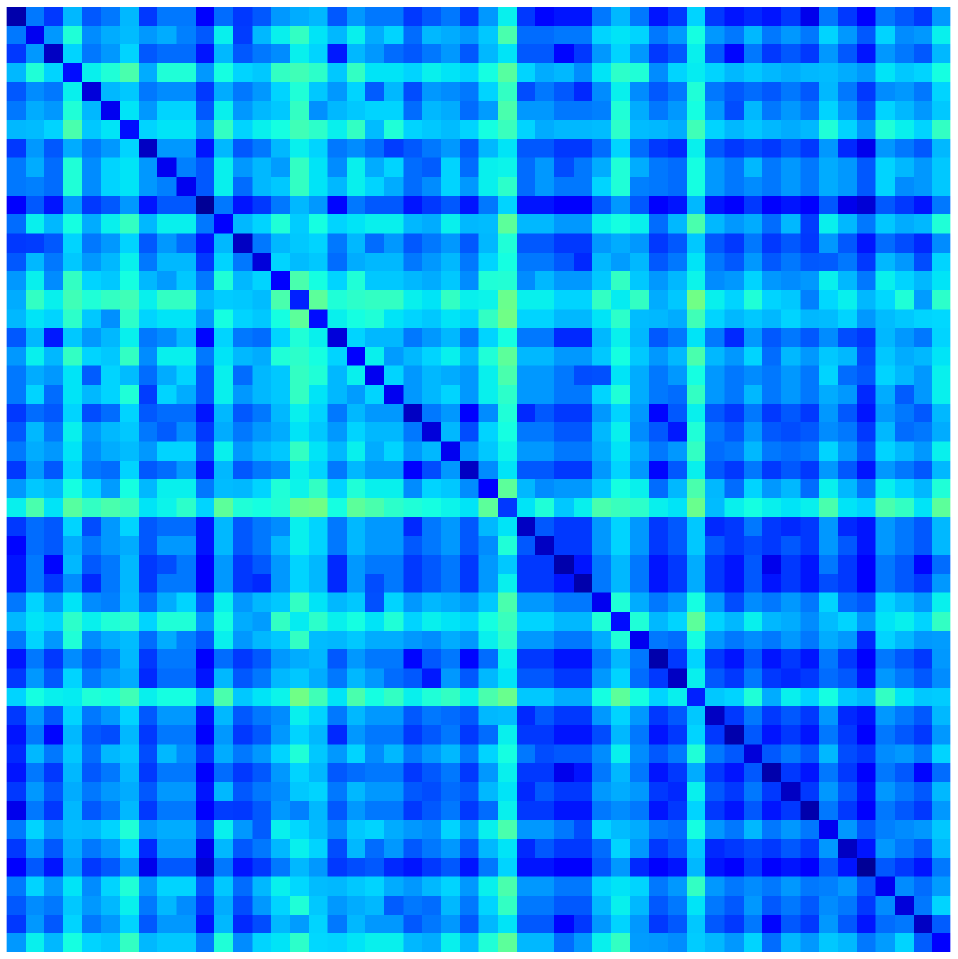}
		\caption{Onlooker network up to a quarter through the simulation with 100 dimensions}\label{fig:100D_heatmap_order_onlooker_from_it_0_until_2485}
	\end{subfigure}
	\hfill
	\begin{subfigure}[h]{0.49\columnwidth}
		\centering
		\includegraphics[width=\textwidth]{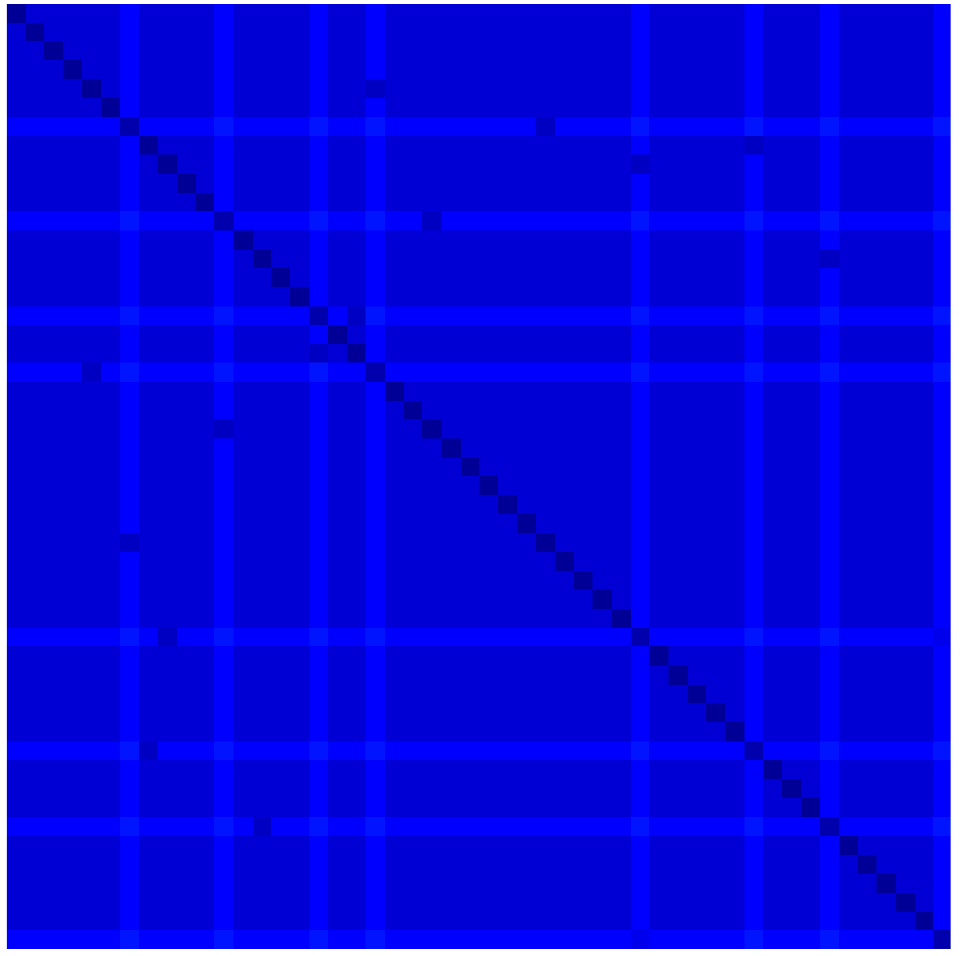}
		\caption{Onlooker network up to halfway of the simulation with 100 dimensions}\label{fig:100D_heatmap_order_onlooker_from_it_0_until_4970}
	\end{subfigure}
\caption{
Influence between bees over time. The color scheme represents the strength of influence between bees from blue as the weakest influence to green as the most substantial influence. Earlier iterations have stronger influence activity, and influence becomes less intense as iterations happen. Onlooker bees display higher levels and more concentrated areas of influence. The employed bees show a little more random influence, which is what we would expect to see due to the nature of the algorithm.
}
\label{fig:onlookermeployeebytime}
\end{figure}

Looking at the time evolution of the employed and onlooker bees in Figure \ref{fig:onlookermeployeebytime}, we observe that a majority of the influence happens at the beginning of the simulation rather than the end. As more iterations happen, it becomes harder for bees to improve and move successfully. Despite having scouts who are randomly assigned to new food sources, the swarm does not seem to have strong enough coordination that allows them to find better regions and avoid worse regions. The algorithm does not provide a strategy for the bees to make wise decisions about where scouts can appear.
Onlooker bees have a slightly more controlled way of exploiting a food source. Because previously successful food sources most influence them, there is a small amount of coordination between them. However, even though they are attracted to the same place, they still do not have enough time to converge in a food source. 

\section{Conclusions}\label{conclusions}

The understanding of swarm-based algorithms is still challenging; few have attempted to explain {\it why} they work. In this paper, we contribute to the area with new insights about the general usefulness of the interaction network framework to evaluate the social interactions of the swarm. We expanded upon Oliveira et al.'s\cite{oliveira2017better} experiment by taking their network-model approach to the PSO and applying it to the ABC. We demonstrate that ABC can be modeled and analyzed by this framework as a multi-layer network in which the aggregated version captures the general behavior within the swarm. We also present that each type of bee or stage provides different influences over time, and the absence of a layer prevents a complete understanding of the algorithm. 

Interestingly, ABC displays a lack of coordination within the swarm which is not present on several configurations of other swarm-based techniques such as PSO \cite{eberhart1995new} and FSS \cite{bastos2009fish}. The non-cooperation is dictated by the excess of dynamic influence through the network supplying diversity on the search space and also on the social behavior. 

Moreover, we can identify that employed and onlooker bees influence the swarm at different levels, but they have similar patterns of influence. The employed bees contribute to the network with a random influence between the bees which adds diversity in the influence patterns. The onlooker bees reveal a higher strength of influence which is expected by the higher probability of successful movements arising from their transition rules. At first glance, scouts may seem unimportant to the process because they generally affect themselves in the multi-layer network. However, without them, the entire ABC algorithm would fail due to lack of diversity. The network that emerges from the two layers (employed and onlooker) is insufficient for ABC to discover other areas of the search space. Note that from Algorithm \ref{alg:ABC} that scouts will assume other roles in the algorithm, so the discovery of new locations is fundamental to the success of ABC. This recognition of different bee patterns expresses the richness of our modeling. 

Lastly, this paper does not argue that our model is the only way that social interactions may be modeled in the ABC. Even previous works on the modeling of PSO do not make such a claim. There may well be other possible representations, but the point of our work is to show that an interaction network modeling is also useful in the case of ABC. 

Future works may attempt to discover alternate ways of modeling the ABC as our goal in this paper was not provide the best way to model the interaction network for ABC, instead, show that using a network-based model can be used to explain what is happening within the algorithm. For instance, one alternative approach to the interaction network could use the food sources as nodes in a bipartite network composed of bees and food sources. Bees move to food sources and interactions could be captured by bees that find themselves in the same location within a time window. Such an approach could highlight the influence of scouts a little more directly. 
Another future work is to analyze how the dynamics of the swarm is affected in constrained optimization problems.

We believe that more work is necessary for the use of interaction network modeling in ABC and other swarm algorithms and that such an approach can unveil similarities between many of the swarm approaches currently in the literature. 


\section*{Acknowledgment}
The authors acknowledge support from National Science Foundation (NSF) grant No. 1560345 and Florida Institute of Technology for providing the computing facilities. Any opinions, findings, and conclusions or recommendations expressed in this material are those of the authors and do not necessarily reflect the views of the NSF. This work used the Extreme Science and Engineering Discovery Environment (XSEDE) Bridges at the Pittsburgh Supercomputing Center through allocation TG-IRI180008, which is supported by National Science Foundation grant number ACI-1548562 \cite{6866038}.

\bibliographystyle{IEEEtran}
\bibliography{main}

\begin{thebibliography}{10}
\providecommand{\url}[1]{#1}
\csname url@samestyle\endcsname
\providecommand{\newblock}{\relax}
\providecommand{\bibinfo}[2]{#2}
\providecommand{\BIBentrySTDinterwordspacing}{\spaceskip=0pt\relax}
\providecommand{\BIBentryALTinterwordstretchfactor}{4}
\providecommand{\BIBentryALTinterwordspacing}{\spaceskip=\fontdimen2\font plus
\BIBentryALTinterwordstretchfactor\fontdimen3\font minus
  \fontdimen4\font\relax}
\providecommand{\BIBforeignlanguage}[2]{{%
\expandafter\ifx\csname l@#1\endcsname\relax
\typeout{** WARNING: IEEEtran.bst: No hyphenation pattern has been}%
\typeout{** loaded for the language `#1'. Using the pattern for}%
\typeout{** the default language instead.}%
\else
\language=\csname l@#1\endcsname
\fi
#2}}
\providecommand{\BIBdecl}{\relax}
\BIBdecl

\bibitem{Bonabeau1999}
E.~Bonabeau, M.~Dorigo, and G.~Theraulaz, \emph{Swarm intelligence: from
  natural to artificial systems}.\hskip 1em plus 0.5em minus 0.4em\relax Oxford
  university press, 1999, no.~1.

\bibitem{Kennedy2001}
J.~Kennedy and R.~C. Eberhart, \emph{{Swarm Intelligence}}.\hskip 1em plus
  0.5em minus 0.4em\relax Morgan Kaufmann Publishers Inc., 2001.

\bibitem{Engelbrecht2006}
A.~P. Engelbrecht, \emph{{Fundamentals of Computational Swarm
  Intelligence}}.\hskip 1em plus 0.5em minus 0.4em\relax John Wiley {\&} Sons,
  2006.

\bibitem{Dorigo2004}
M.~Dorigo and T.~St{\"u}tzle, \emph{{Ant Colony Optimization}}.\hskip 1em plus
  0.5em minus 0.4em\relax The MIT Press, 2004.

\bibitem{karaboga2008performance}
D.~Karaboga and B.~Basturk, ``On the performance of artificial bee colony (abc)
  algorithm,'' \emph{Applied soft computing}, vol.~8, no.~1, pp. 687--697,
  2008.

\bibitem{yang2010firefly}
X.-S. Yang, ``Firefly algorithm, l{\'{e}}vy flights and global optimization,''
  in \emph{Research and Development in Intelligent Systems {XXVI}}, 2009, pp.
  209--218.

\bibitem{bastos2008novel}
C.~J. Bastos~Filho, F.~B. de~Lima~Neto, A.~J. Lins, A.~I. Nascimento, and M.~P.
  Lima, ``A novel search algorithm based on fish school behavior,'' in
  \emph{Systems, Man and Cybernetics, 2008. SMC 2008. IEEE International
  Conference on}.\hskip 1em plus 0.5em minus 0.4em\relax IEEE, 2008, pp.
  2646--2651.

\bibitem{oliveira2018unveiling}
M.~Oliveira, D.~Pinheiro, M.~Macedo, C.~Bastos-Filho, and R.~Menezes,
  ``Unveiling swarm intelligence with network science---the metaphor
  explained,'' \emph{arXiv preprint arXiv:1811.03539}, 2018.

\bibitem{bonabeau2000inspiration}
E.~Bonabeau, M.~Dorigo, and G.~Theraulaz, ``Inspiration for optimization from
  social insect behaviour,'' \emph{Nature}, vol. 406, no. 6791, p.~39, 2000.

\bibitem{Sorensen2015}
K.~S{\"{o}}rensen, ``Metaheuristics-the metaphor exposed,'' \emph{International
  Transactions in Operational Research}, vol.~22, no.~1, pp. 3--18, 2013.

\bibitem{bratton2007understanding}
D.~Bratton and T.~Blackwell, ``Understanding particle swarms through
  simplification: a study of recombinant pso,'' in \emph{Proceedings of the 9th
  annual conference companion on Genetic and evolutionary computation}.\hskip
  1em plus 0.5em minus 0.4em\relax ACM, 2007, pp. 2621--2628.

\bibitem{eberhart1995new}
R.~Eberhart and J.~Kennedy, ``A new optimizer using particle swarm theory,'' in
  \emph{Micro Machine and Human Science, 1995. MHS'95., Proceedings of the
  Sixth International Symposium on}.\hskip 1em plus 0.5em minus 0.4em\relax
  IEEE, 1995, pp. 39--43.

\bibitem{complenet2013}
M.~Oliveira, C.~J.~A. Bastos-Filho, and R.~Menezes, ``Assessing particle swarm
  optimizers using network science metrics,'' in \emph{Complex Networks {IV}},
  ser. Studies in Computational Intelligence, 2013, vol. 476, pp. 173--184.

\bibitem{oliveira2014towards}
M.~Oliveira, C.~J. Bastos-Filho, and R.~Menezes, ``Towards a network-based
  approach to analyze particle swarm optimizers,'' in \emph{Swarm Intelligence
  (SIS), 2014 IEEE Symposium on}.\hskip 1em plus 0.5em minus 0.4em\relax IEEE,
  2014, pp. 1--8.

\bibitem{Oliveira2015}
M.~Oliveira, C.~J.~A. Bastos-Filho, and R.~Menezes, ``Using network science to
  assess particle swarm optimizers,'' \emph{Social Network Analysis and
  Mining}, vol.~5, no.~1, pp. 1--13, 2015.

\bibitem{oliveira2016communication}
M.~Oliveira, D.~Pinheiro, B.~Andrade, C.~Bastos-Filho, and R.~Menezes,
  ``Communication diversity in particle swarm optimizers,'' in
  \emph{International Conference on Swarm Intelligence}.\hskip 1em plus 0.5em
  minus 0.4em\relax Springer, 2016, pp. 77--88.

\bibitem{oliveira2017better}
M.~Oliveira, D.~Pinheiro, M.~Macedo, C.~Bastos-Filho, and R.~Menezes, ``Better
  exploration-exploitation pace, better swarm: Examining the social
  interactions,'' in \emph{Computational Intelligence (LA-CCI), 2017 IEEE Latin
  American Conference on}.\hskip 1em plus 0.5em minus 0.4em\relax IEEE, 2017,
  pp. 1--6.

\bibitem{karaboga2005idea}
D.~Karaboga, ``An idea based on honey bee swarm for numerical optimization,''
  Technical report-tr06, Erciyes university, engineering faculty, computer
  engineering department, Tech. Rep., 2005.

\bibitem{bastos2009fish}
C.~J. Bastos~Filho, F.~B. de~Lima~Neto, A.~J. Lins, A.~I. Nascimento, and M.~P.
  Lima, ``Fish school search,'' in \emph{Nature-inspired algorithms for
  optimisation}.\hskip 1em plus 0.5em minus 0.4em\relax Springer, 2009, pp.
  261--277.

\bibitem{6866038}
J.~Towns, T.~Cockerill, M.~Dahan, I.~Foster, K.~Gaither, A.~Grimshaw,
  V.~Hazlewood, S.~Lathrop, D.~Lifka, G.~D. Peterson, R.~Roskies, J.~R. Scott,
  and N.~Wilkins-Diehr, ``Xsede: Accelerating scientific discovery,''
  \emph{Computing in Science and Engineering}, vol.~16, no.~5, pp. 62--74,
  Sept.-Oct. 2014.

\end{thebibliography}

\end{document}